\pdfoutput=1

\documentclass[11pt]{article}

\usepackage{ACL2023}
\usepackage{multicol}
\usepackage{multirow}
\usepackage{amsfonts}
\usepackage{graphicx} 
\graphicspath{ {./images/} }
\usepackage{graphics} 
\usepackage{adjustbox}

\usepackage{times}
\usepackage{latexsym}

\usepackage[T1]{fontenc}

\usepackage[utf8]{inputenc}

\usepackage{microtype}

\usepackage{inconsolata}
\usepackage{amsmath}
\usepackage{booktabs}
\usepackage{makecell}
\usepackage{multirow}

\title{Clinical Note Owns its Hierarchy: Multi-Level Hypergraph \\Neural Networks for Patient-Level Representation Learning}

\author{Nayeon Kim\textsuperscript{\rm 1*}, Yinhua Piao\textsuperscript{\rm 2*}, \and Sun Kim\textsuperscript{\rm 1,2,3,4} \\
        \textsuperscript{\rm 1} Interdisciplinary Program in Artificial Intelligence, Seoul National University \\ 
        \textsuperscript{\rm 2} Department of Computer Science and Engineering, Seoul National University \\ 
        \textsuperscript{\rm 3} Institute of Computer Technology, Seoul National University\\
        \textsuperscript{\rm 4} AIGENDRUG Co., Ltd.\\
        \texttt{$\{$ny\_1031, 2018-27910, sunkim.bioinfo$\}$@snu.ac.kr} }

\begin{document}
\maketitle

\def\thefootnote{*}\footnotetext{These authors contributed equally to this work.}\def\thefootnote{\arabic{footnote}} 

\begin{abstract}

Leveraging knowledge from electronic health records (EHRs) to predict a patient's condition is essential to the effective delivery of appropriate care.
Clinical notes of patient EHRs contain valuable information from healthcare professionals,
but have been underused due to their difficult contents and complex hierarchies.
Recently, hypergraph-based methods have been proposed for document classifications. 
Directly adopting existing hypergraph methods on clinical notes cannot sufficiently utilize the hierarchy information of the patient, which can degrade clinical semantic information by (1) \textit{frequent neutral words} and (2) \textit{hierarchies with imbalanced distribution}.
Thus, we propose a taxonomy-aware multi-level hypergraph neural network (TM-HGNN), where multi-level hypergraphs assemble useful neutral words with rare keywords via note and taxonomy level hyperedges to retain the clinical semantic information. The constructed patient hypergraphs are fed into hierarchical message passing layers for learning more balanced multi-level knowledge at the note and taxonomy levels. 
We validate the effectiveness of TM-HGNN by conducting extensive experiments with MIMIC-III dataset on benchmark in-hospital-mortality prediction.\footnote{Our codes and models are publicly available at: \\  \url{https://github.com/ny1031/TM-HGNN}}


\end{abstract}

\section{Introduction}
\label{introduction}
\begin{figure}[t]
\centering

\includegraphics[width=1\columnwidth]{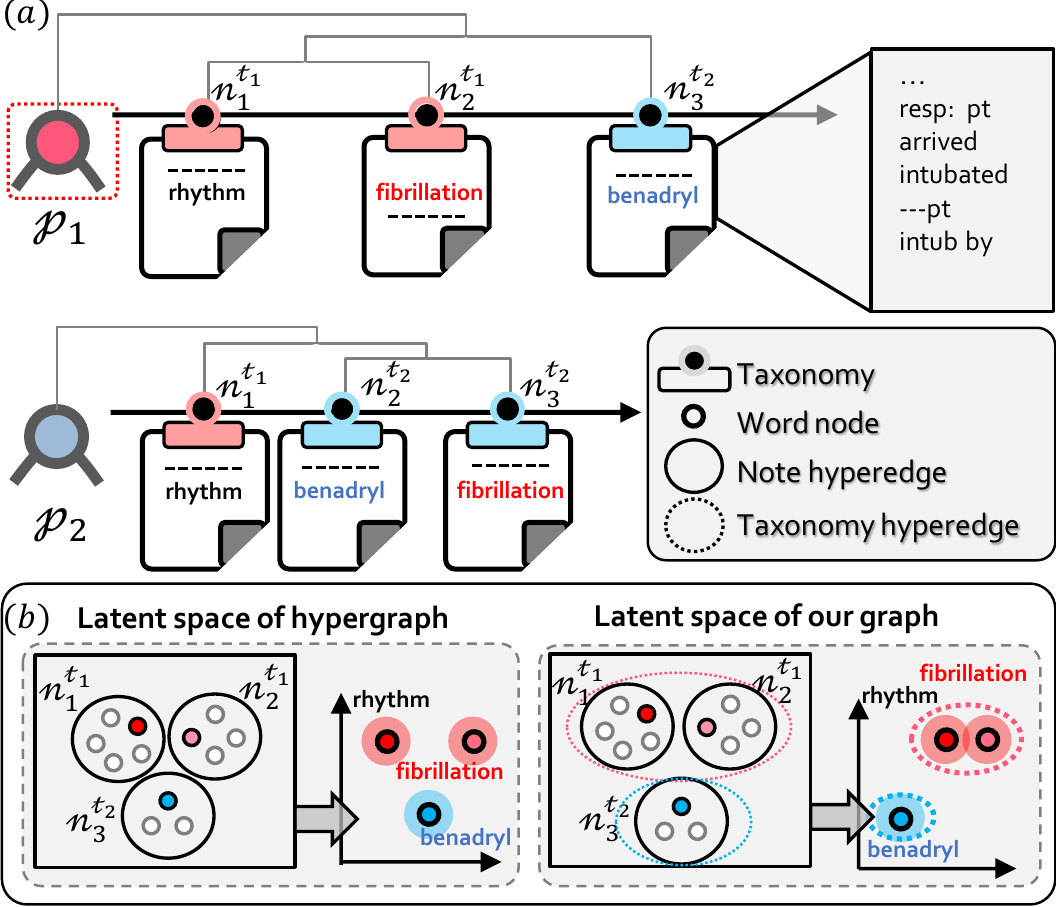} 
\caption{(a) Examples of patient clinical notes with difficult contents (e.g. jargons and abbreviations) and complex structures. Patient $p_1$ owns notes of radiology taxonomy (pink) and nursing taxonomy (blue). (b) Differences between existing hypergraphs and our proposed multi-level hypergraphs.}
\label{fig1}
\end{figure}
\begin{figure}[t]
\centering
\includegraphics[width=1\columnwidth]{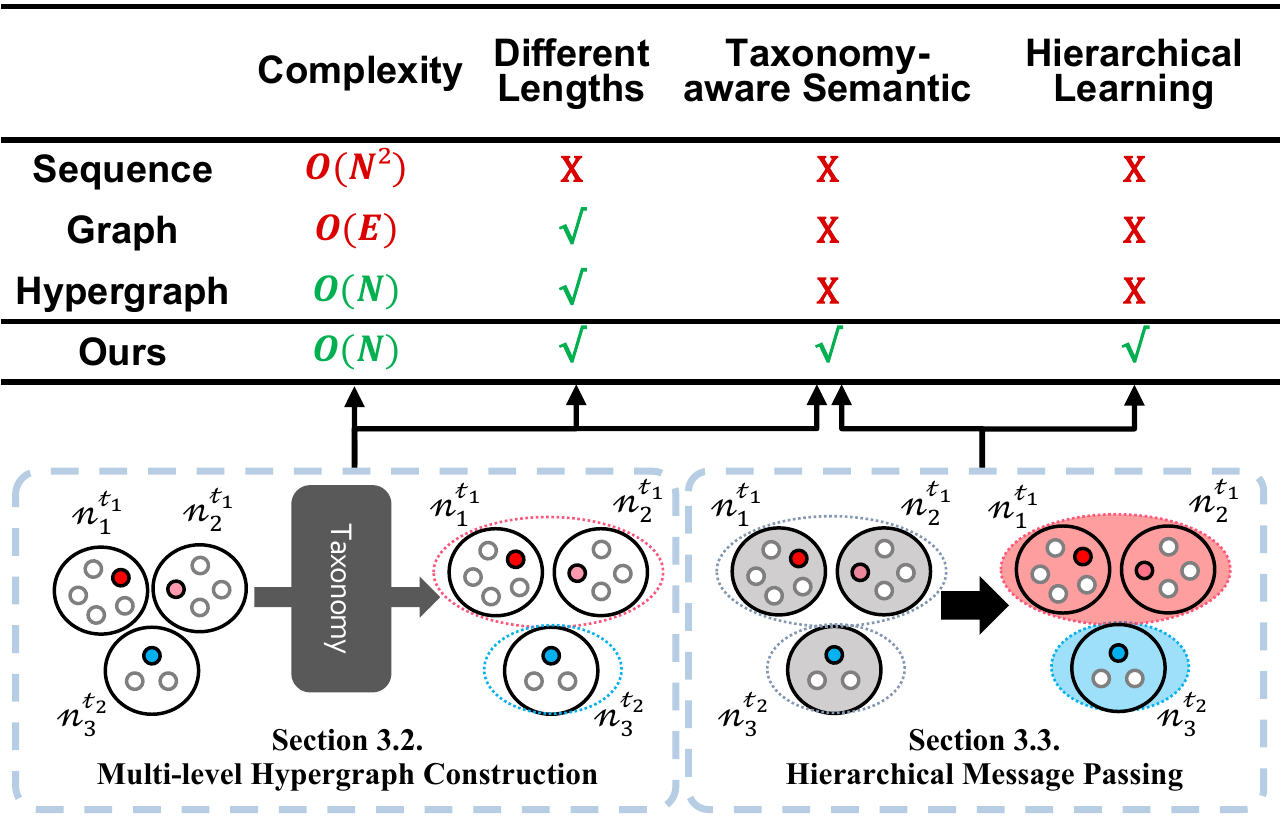} 
\caption{Advantages of the proposed model, compared to sequence, graph and hypergraph based models. $N$ and $E$ denote the number of nodes and edges respectively. We address issues of complexity and different lengths by adopting the hypergraph to represent each patient. Our model retains semantic information by constructing multi-level hypergraph (Section \ref{sec: multilevel}), and hierarchical message passing layers (Section \ref{sec: hierarchical}) are proposed for balancing multi-level knowledge for patient representation learning.}
\label{summary}
\end{figure}

With improvement in healthcare technologies, electronic health records (EHRs) are being used to monitor intensive care units (ICUs) in hospitals. Since it is crucial to schedule appropriate treatments for patients in ICUs, 
there are many prognostic models that use EHRs to address related tasks, such as in-hospital mortality prediction.
EHRs consist of three types of data; structured, semi-structured, and unstructured. Clinical notes, which are unstructured data, contain valuable comments or summary of the patient's condition written by medical professionals (doctors, nurses, etc.). 
However, compared to structured data, clinical notes have been underutilized in previous studies due to the difficult-to-understand contents and the complex hierarchies (Figure \ref{fig1}(a)).
Transformer-based \cite{vaswani2017attention} methods like ClinicalBERT \cite{alsentzer2019publicly, huang2019clinicalbert,huang2020clinical} have been proposed to pre-train on large-scale corpus from similar domains, and fine-tune on the clinical notes through transfer learning.
While Transformer-based methods can effectively detect distant words compared to other sequence-based methods like convolutional neural networks \cite{kim-2014-convolutional, zhang2015character} and recurrent neural networks \cite{mikolov2010recurrent, tai2015improved, liu2016recurrent}, there are still limitations of increasing computational complexity for long clinical notes (Figure \ref{summary}).

Recently, with the remarkable success of the graph neural networks (GNNs) \cite{kipf2017semisupervised, velivckovic2018graph, brody2021attentive}, graph-based document classification methods have been proposed \cite{yao2019graph, huang2019text} that can capture long range word dependencies 
and can be adapted to documents with different and irregular lengths.
Some methods build word co-occurrence graphs by sliding fixed-size windows to model pairwise interactions between words \cite{zhang2020every, piao2022sparse, wang2022induct}. However, the density of the graph increases as the document becomes longer. Besides, there are also some methods apply hypergraph for document classification \cite{ding2020more, zhang2022hegel}, which can alleviate the high density of the document graphs and extract high-order structural information of the documents. 

Adopting hypergraphs can reduce burden for managing long documents with irregular lengths, but additional issues remain when dealing with clinical notes: \textbf{\textit{(1) Neutral words deteriorate clinical semantic information.}} 
In long clinical notes, there are many frequently written neutral words (e.g. "\textit{rhythm}") that do not directly represent the patient's condition. 
Most of the previous methods treat all words equally at the learning stage, which may result in dominance of frequent neutral words, and negligence of rare keywords that are directly related to the patient's condition.
Meanwhile, the neutral word can occasionally augment information of rare keywords, depending on the intra-taxonomy context. 
Taxonomy represents the category of the clinical notes, where implicit semantic meaning of the words can differ. 
For example, "\textit{rhythm}" occurred with "\textit{fibrillation}" in \textit{ECG} taxonomy can represent serious cardiac disorder of a patient, but when "\textit{rhythm}" is written with "\textit{benadryl}" in \textit{Nursing} taxonomy, it can hardly represent the serious condition. 
Therefore, assembling intra-taxonomy related words can leverage "\textit{useful}" neutral words with rare keywords to jointly augment the 
clinical semantic information,
which implies the necessity of introducing taxonomy-level hyperedges. (2) \textbf{\textit{Imbalanced distribution of multi-level hyperedges.}} There are a small number of taxonomies compared to notes for each patient. As a result, when taxonomy-level and note-level information are learned simultaneously, note-level information can obscure taxonomy-level information. To learn more balanced multi-level information of the clinical notes, an effective way for learning the multi-level hypergraphs with imbalanced distributed hyperedges is required. 

To address the above issues, we propose TM-HGNN (Taxonomy-aware Multi-level HyperGraph Neural Networks), which can effectively and efficiently utilize the multi-level high-order semantic information for patient representation learning. 
Specifically, we adopt patient-level hypergraphs to manage highly unstructured and long clinical notes and define multi-level hyperedges, i.e., note-level and taxonomy-level hyperedges. Moreover, we conduct the hierarchical message passing from note-level to taxonomy-level hyperedges using edge-masking. 
To hierarchically learn word embeddings without mixture of information between note and taxonomy, note and taxonomy hyperedges are disconnected.
Note-level word embeddings are learned only with intra-note local information. The following taxonomy-level propagation introduce clinical semantic information by assembling the intra-taxonomy words and separating inter-taxonomy words for better patient-level representation learning. The contributions of this article can be summarized as follows (Figure \ref{summary}):
\begin{itemize}
\item To address issue 1, we construct multi-level hypergraphs for patient-level representation learning, which can assemble "\textit{useful}" neutral word with rare keyword via note and taxonomy level hyperedges to retain the clinical semantic information.
\item To address issue 2, we propose hierarchical message passing layers for the constructed graphs with imbalanced hyperedges, which can learn more balanced multi-level knowledge for patient-level representation learning. 
\item We conduct experiments with MIMIC-III clinical notes on benchmark in-hospital-mortality task. The experimental results demonstrate the effectiveness of our approach.
\end{itemize}

\section{Related Work}
\subsection{Models for Clinical Data}
With the promising potential of managing medical data, four benchmark tasks were proposed by \citet{harutyunyan2019multitask} for MIMIC-III (Medical Information Mart for Intensive Care-III) \cite{johnson2016mimic} clinical dataset. Most of the previous works with MIMIC-III dataset focus on the structured data (e.g. vital signals with time-series) for prognostic prediction tasks \cite{choi2016retain,shang2019gamenet} or utilize clinical notes combined with time-series data \cite{khadanga2019using, deznabi2021predicting}. Recently, there are approaches focused on clinical notes, adopting pre-trained models such as BERT-based \cite{alsentzer2019publicly, huang2019clinicalbert, golmaei2021deepnote,naik2022literature} and XLNet-based \cite{huang2020clinical} 
or utilizing contextualized phenotypic features extracted from clinical notes \cite{zhang2022clinical}.

\subsection{Graph Neural Networks for Document Classification}
Graph neural networks \cite{kipf2017semisupervised, velivckovic2018graph, brody2021attentive} have achieved remarkable success in various deep learning tasks, including text classification. Initially, transductive graphs have been applied to documents, such as TextGCN \cite{yao2019graph}. Transductive models have to be retrained for every renewal of the data, which is inefficient and hard to generalize \cite{yao2019graph,huang2019text}. For inductive document graph learning, word co-occurrence graphs initialize nodes with word embeddings and exploit pairwise interactions between words. TextING \cite{zhang2020every} employs the gated graph neural networks for document-level graph learning. Following TextGCN \cite{yao2019graph} which applies graph convolutional networks (GCNs) \cite{kipf2017semisupervised} in transductive level corpus graph, InducT-GCN \cite{wang2022induct} applies GCNs in inductive level where unseen documents are allowed to use. TextSSL \cite{piao2022sparse} captures both local and global structural information within graphs.

However, the density of word co-occurrence graph increases as the document becomes longer, since the fixed-sized sliding windows are used to capture local pairwise edges. In case of hypergraph neural networks, hyperedges connect multiple number of nodes instead of connecting words to words by edges, which alleviates the high density of the text graphs. HyperGAT \cite{ding2020more} proposes document-level hypergraphs with hyperedges containing sequential and semantic information. HEGEL \cite{zhang2022hegel} applies Transformer-like \cite{vaswani2017attention} multi-head attention to capture high-order cross-sentence relations for effective summarization of long documents. According to the reduced computational complexity for long documents (Figure \ref{summary}), we adopt hypergraphs to represent patient-level EHRs with clinical notes. Considering issues of existing hypergraph-based methods (Figure \ref{summary}), we construct multi-level hypergraphs at note-level and taxonomy-level for each patient. The constructed graphs are fed into hierarchical message passing layers to capture rich hierarchical information of the clinical notes, which can augment semantic information for patient representation learning.






\section{Method}

\begin{figure*}[t]
\includegraphics[width=1\textwidth]{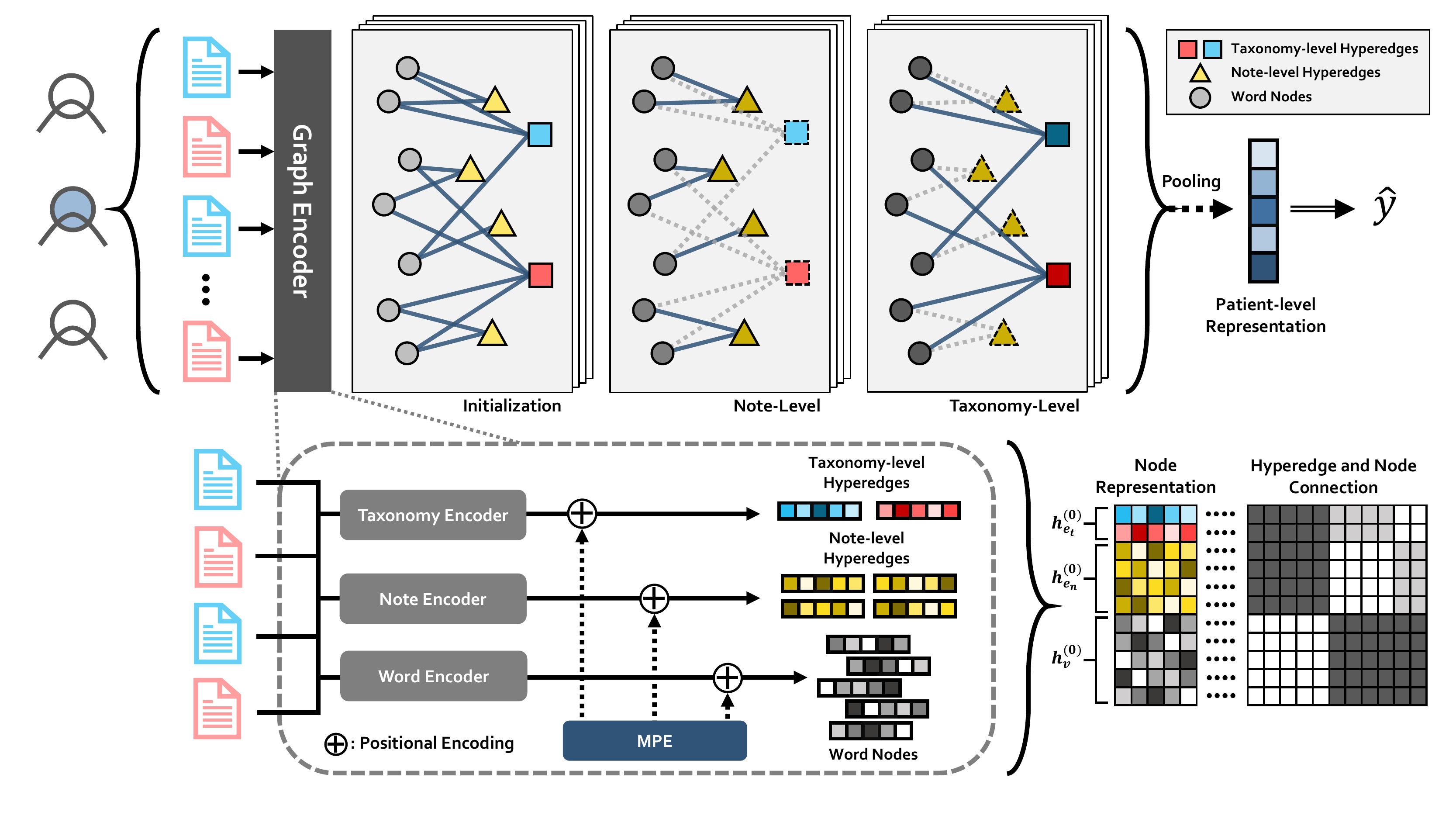} 
\caption{Overview of the proposed TM-HGNN. Taxonomy-aware multi-level hypergraphs are fed into the model for hierarchical message passing. \(\hat{y}\) denotes the patient-level prediction.}
\label{model_structure}
\end{figure*}


\subsection{Problem Definition}

Our task is to predict in-hospital-mortality for each patient using a set of clinical notes. Given a patient $p\in\mathcal{P}$ with in-hospital-mortality label $y\in\mathcal{Y}$,  patient $p$ owns a list of clinical notes $\mathcal{N}_{p}=[n_1^{t_1}, ..., n_j^{t_k}, ...]$, and each clinical note $n^{t}\in\mathcal{N}_p$ with taxonomy $t\in\mathcal{T}_p$ contains a sequence of words $\mathcal{W}_{n^t}=[w_1^{n^{t}}, ..., w_i^{n^t}, ...]$, where $j$, $k$ and $i$ denote the index of clinical note $n$, taxonomy $t$ and word $w$ of patient $p$. The set of taxonomies can be represented by $\mathcal{T}=\{t_1, t_2, ..., t_k, ...\}$. 

Our goal is to construct individual multi-level hypergraphs $\mathcal{G}_p$ for each patient $p$ and learn patient-level representation $\mathcal{G}_p$ with the multi-level knowledge  by hierarchical message passing layers for in-hospital-mortality prediction tasks. Since our model is trained by inductive learning, patient $p$ is omitted throughout the paper. 

\subsection{Multi-Level Hypergraph Construction}\label{sec: multilevel}

We construct multi-level hypergraphs for patient-level representation learning, which can address the issues that are mentioned in introduction \ref{introduction}.
A hypergraph $\mathcal{G}^*=(\mathcal{V}, \mathcal{E})$ consists of a set of nodes $\mathcal{V}$ and hyperedges $\mathcal{E}$ where multiple nodes can be connected to single hyperedge $e\in\mathcal{E}$. 
A multi-level hypergraph $\mathcal{G}=\{\mathcal{V}, \{\mathcal{E}_\mathcal{N}\cup\mathcal{E}_\mathcal{T}\}\}$ is constructed from patient's clinical notes, where $\mathcal{E_N}$ and $\mathcal{E_T}$ denote note-level and taxonomy-level hyperedges, respectively. A word node $v$ exists in note $n$ with the taxonomy of $t$ can be represented by $\{v\in{n},n\in{t}\}$. A note-level hyperedge is denoted as $e_n$, and a taxonomy-level hyperedge is denoted as $e_t$.

\paragraph{Multi-level Positional Encoding} 
There are three types of entries in the multi-level hypergraph $\mathcal{G}$, such as word nodes $\mathcal{V}$, note-level hyperedges $\mathcal{E}_\mathcal{N}$ and taxonomy-level hyperedges $\mathcal{E}_\mathcal{T}$. 
To distinguish these entries, we propose multi-level positional encoding to introduce more domain-specific meta-information to the hypergraph $\mathcal{G}$.
The function of multi-level positional encoding $\scriptstyle \mathbf{MPE}(\cdot)$ can be defined as:

\small
\begin{equation}
\mathbf{MPE}(x)=[\tau(x), \mathcal{I}_\mathcal{W}(x), \mathcal{I}_\mathcal{N}(x), \mathcal{I}_\mathcal{T}(x)]
\end{equation}
\normalsize

where entry $x \in \{\mathcal{V}, \mathcal{E}_\mathcal{N}, \mathcal{E}_\mathcal{T}\}$, and function $\tau: x \mapsto \{0, 1, 2\}$ maps entry $x$ to a single type among nodes, note-level and taxonomy-level hyperedges. Functions $\mathcal{I}_{\mathcal{W}}(\cdot)$, $\mathcal{I}_{\mathcal{N}}(\cdot)$, and $\mathcal{I}_{\mathcal{T}}(\cdot)$ maps entry $x$ to positions in the word, note and taxonomy-level, respectively. To initialize embedding of node $v$, we concatenate embedding $\scriptstyle \mathbf{MPE}(v)$ from multi-level position encoding and word2vec \cite{mikolov2010recurrent} pre-trained embedding $\mathbf{z}_v$. 
Since shallow word embeddings are widely used to initialize node embeddings in graph-based document representation \cite{10.1145/3375395.3387641}, we use word2vec \cite{mikolov2010recurrent} embedding.
A word node embedding $\mathbf{h}^{(0)}_v$ is constructed as follows:  

\small
\begin{equation}
\mathbf{h}^{(0)}_v = \mathbf{MPE}(v) \oplus \mathbf{z}_v,
\end{equation}
\normalsize

\vspace{0.04cm}

where $\oplus$ denotes concatenation function.

\subsubsection{Hyperedge Construction}
To extract multi-level information of patient-level representation using clinical notes, we construct patient hypergraphs with two types of hyperedges, one at the note-level hyperedge $\mathcal{E_N}$ and the other at the taxonomy-level hyperedge $\mathcal{E_T}$. A word node $v$ in note $n$ with taxonomy $t$ is assigned to one note-level hyperedge $e_n$ and one taxonomy-level hyperedge $e_t$, which can be defined as:

\small
\begin{equation}
\mathcal{E}(v)=\{e_n,e_t|v\in{n},{n\in{t}}\}
\end{equation}
\normalsize


\paragraph{Note-level Hyperedges} 
We adopt linear embedding function $f_n$ and obtain the index embedding using $\mathcal{I}_\mathcal{N}(n)$. To preserve time-dependent sequential information of clinical note $n$, we simply add time information $\mathbf{t}(n)$ to the embedding. Then initial embedding of note-level hyperedge $h^{(0)}_{e_n}$ with $\scriptstyle \mathbf{MPE}(\cdot)$ can be defined as:

\small
\begin{equation}
\mathbf{h}^{(0)}_{e_n} = \mathbf{MPE}(n) \oplus f^\theta_{n}\bigl(\mathcal{I}_\mathcal{N}(n),\mathbf{t}(n)\bigl),\\
\end{equation}
\normalsize

where $\theta\in\mathbb{R}^{d\times{d}}$ denotes the parameter matrix of function $f_n$. Notably, we set the value of word index   $\mathcal{I}_\mathcal{W}(n)$ as -1 since the note $n$ represents higher level information than word $v$.

\paragraph{Taxonomy-level Hyperedges} 
Taxonomy-level hyperedges $e_t$ are constructed by taxonomy index $\mathcal{I}_\mathcal{T}(t)$ through linear layers $f_t$ concatenated with $\scriptstyle \mathbf{MPE}(\cdot)$ function, which can be defined as:

\small
\begin{equation}
\mathbf{h}^{(0)}_{e_t} = \mathbf{MPE}(t) \oplus f^{\theta}_{t}\bigl(\mathcal{I}_\mathcal{T}(t)\bigl),\\
\end{equation}
\normalsize

where $\theta\in\mathbb{R}^{d\times{d}}$ denotes the parameter matrix of function $f_t$. Like note-level hyperedge, we set $\mathcal{I}_\mathcal{W}(t)$ and $\mathcal{I}_\mathcal{N}(t)$ as -1 since the level of taxonomy $t$ is higher than the levels of note and word.

\subsection{Hierarchical Message Passing}\label{sec: hierarchical}
To leverage the characteristics of two types of hyperedges, we propose a hierarchical hypergraph convolutional networks, composed of three layers that allow message passing from different types of hyperedges. 
In general, we define message passing functions for nodes and hyperedges as follows: 

\small
\begin{equation}
{\mathcal{F}}_{\mathcal{W}}(\mathrm{h}, \mathcal{E}, \theta) = \sigma \biggl( \theta\biggl(\sum_{\substack{u\in\mathcal{E}(v)}}\frac{1}{\sqrt{\hat{d_{v}}}\sqrt{\hat{d_u}}}\mathrm{h}_u\biggl)\biggl), \\
\end{equation}
\begin{equation}
{\mathcal{F}}_{\mathcal{\tau}}(\mathrm{h}, \mathcal{V}^\tau, \theta) = \sigma \biggl( \theta\biggl(\sum_{\substack{z\in {\mathcal{V}^\tau(e)}}}\frac{1}{\sqrt{\hat{d_{e}}}\sqrt{\hat{d_z}}}\mathrm{h}_z\biggl)\biggl), \\
\end{equation}
\normalsize
\vspace{0.03cm}

where \(\mathcal{F}_\mathcal{W}\) denotes message passing function for word nodes and $\mathcal{F}_\tau$ denotes message passing function for hyperedges with type $\tau\in\{1,2\}$, i.e., note-level hyperedges and taxonomy-level hyperedges, respectively.
Function $\mathcal{F}_\mathcal{W}$ updates word node embedding $\mathrm{h}_v$ by aggregating embeddings of connected hyperedges $\mathcal{E}(v)$ . 
Function $\mathcal{F}_\tau$ updates hyperedge embedding $\mathrm{h}_e$ by aggregating embeddings of connected word nodes $\mathcal{V}^\tau(e)$. \(\sigma\) is the nonlinear activation function such as \(\mathrm{ReLU}\), $\theta\in\mathbb{R}^{d\times{d}}$ is the weight matrix with dimension $d$ which can be differently assinged and learned at multiple levels. 

Then we can leverage these defined functions to conduct hierarchical message passing learning at the note level and at the taxonomy level.
\begin{table}[t]
\centering
\begin{tabular}{lll}
\toprule
& Statistics \\
\hline
\# of patients & 17,927 \\
\# of ICU stays & 21,013 \\
\hline
\# of in-hospital survival & 18,231 \\ 
\# of in-hospital mortality & 2,679 \\
\hline
\# of notes per ICU stay & 13.29 (7.84) \\
\# of words per ICU stay & 1,385.62 (1,079.57)\\
\# of words per note & 104.25 (66.82)\\
\# of words per taxonomy & 474.75 (531.42)\\
\bottomrule
\end{tabular}
\caption{Statistics of the MIMIC-III clinical notes. Averaged numbers are reported with standard deviation.}
\label{table1}
\end{table}

\paragraph{Initialization Layer}
Due to the complex structure of the clinical notes, the initial multi-level hypergraph constructed for each patient has a large variance. To prevent falling into local optima in advance, we first use an initialization layer to pre-train the entries of hypergraphs by learning the entire patient graph structure.
In this layer, message passing functions are applied to all word nodes $v\in\mathcal{V}$ and hyperedges $e\in\mathcal{E_I}=\{\mathcal{E_N}\cup\mathcal{E_T}\}$. Thus, embeddings of node $v$, hyperedges $e_n$ and $e_t$ at both levels can be defined as: 

\small
\begin{equation}
h_I(v) = \mathcal{F}_{\mathcal{W}}\bigl(h^{(0)}_v, \mathcal{E_I}(v), \theta_{I}\bigl),\\
\end{equation}
\vspace{0.0003cm}
\begin{equation}
h_I(e_n) = \mathcal{F}_{\tau}\bigl(h^{(0)}_{e_n}, \mathcal{V}^{\tau}(e_n),\theta_I\bigl),  \tau=1\\
\end{equation}
\vspace{0.0003cm}
\begin{equation}
h_I{(e_t)} = \mathcal{F}_{\tau}\bigl(h^{(0)}_{e_t}, \mathcal{V}^{\tau}(e_t), \theta_I\bigl), \tau=2 \\
\end{equation}
\normalsize

\paragraph{Note-level Message Passing Layer}
Then we apply note-level message passing layer on hypergraphs with only word nodes $v\in\mathcal{V}$ and note-level hyperedges $e_n\in\mathcal{E_N}$, and the taxonomy-level hyperedges are masked during message passing. In this layer, the word nodes can only interact with note-level hyperedges, which can learn the intra-note local information.

\small
\begin{equation}
h_N(v) = \mathcal{F}_{\mathcal{W}}\bigl(h_I(v), \mathcal{E_N}(v), \theta_N\bigl),
\end{equation}
\vspace{0.0003cm}
\begin{equation}
h_N(e_n) = \mathcal{F}_{\tau}\bigl(h_I(e_n), \mathcal{V}^{\tau}(e_n), \theta_N\bigl), \tau=1,
\end{equation}
\vspace{0.0003cm}
\begin{equation}
h_N(e_t) = h_I(e_t)
\end{equation}
\normalsize

\paragraph{Taxonomy-level Message Passing Layer}
The last layer is the taxonomy-level message passing layer, where all word nodes $v\in\mathcal{V}$ and taxonomy-level hyperedges $e_t\in\mathcal{E_T}$ can be updated. In this layer, we block the hyperedges at the note level. The node representations with note-level information are fused with taxonomy information via taxonomy-level hyperedges, which can assemble the intra-taxonomy related words to augment semantic information.

\small
\begin{equation}
h_T(v) = \mathcal{F}_{\mathcal{W}}\bigl(h_N(v), \mathcal{E_T}(v), \theta_T\bigl),
\end{equation}
\begin{equation}
h_T(e_n) = h_N(e_n),
\end{equation}
\begin{equation}
h_T(e_t) = \mathcal{F}_{\tau}\bigl(h_N(e_t), \mathcal{V}^{\tau}(e_t),\theta_T\bigl), \tau=2 \\
\end{equation}
\normalsize

\subsubsection{Patient-Level Hypergraph Classification}
After all aforementioned hierarchical message passing layers, node and hyperedge embeddings $h_T(v),h_T(e_n),h_T(e_t)\in\mathbf{H}_T$ follow mean-pooling operation which summarizes patient-level embedding \(z\), which is finally fed into $\mathrm{sigmoid}$ operation as follows:

\small
\begin{equation}
\hat{y} = \mathrm{sigmoid}(z)
\label{10}
\end{equation}
\normalsize

where \(\hat{y}\) denotes the probability of the predicted label for in-hospital-mortality of the patient. The loss function for patient-level classification is defined as the binary cross-entropy loss:

\small
\begin{equation}
\mathcal{L} = -\left(y\times \log\hat{y}+(1-y)\times\log(1-\hat{y})\right)
\label{10}
\end{equation}
\normalsize

where \(y\) denotes the true label for in-hospital-mortality. The proposed network, TM-HGNN, can be trained by minimizing the loss function.

\section{Experimental Settings}
\subsection{Dataset}
We use clinical notes from the Medical Information Mart for Intensive Care III (MIMIC-III) \cite{johnson2016mimic} dataset, which are written within 48 hours from the ICU admission. For quantitative evaluation, we follow \citeposs{harutyunyan2019multitask} benchmark setup for data pre-processing and train/test splits, then randomly divide 20\% of train set as validation set. All patients without any notes are dropped during the data preparation. To prevent overfitting into exceptionally long clinical notes for a single patient, we set the maximum number of notes per patient into 30 from the first admission. Table \ref{table1} shows the statistics of pre-processed MIMIC-III clinical note dataset for our experiments. We select top six taxonomies for experiments, since the number of notes assigned to each taxonomy differs in a wide range (Appendix \ref{train_stats} Table \ref{tax_stat}). In addition, we select two chronic diseases, hypertension and diabetes, to compare prediction results for patients with each disease.

\subsection{Compared Methods}
In our experiments, the compared baseline methods for end-to-end training are as follows:
\begin{itemize}
\item Word-based methods: word2vec \cite{NIPS2013_9aa42b31} with multi-layer perceptron classifier, and FastText \cite{joulin2017bag}.
\item Sequence-based methods: TextCNN \cite{kim-2014-convolutional}, Bi-LSTM \cite{hochreiter1997long}, and Bi-LSTM with additional attention layer \cite{zhou2016attention}.
\item Graph-based methods: TextING \cite{zhang2020every}, InducT-GCN \cite{wang2022induct}, and HyperGAT \cite{ding2020more}. In particular, HyperGAT represents hypergraph-based method, and the other graph-based methods employ word co-occurrence graphs.
\end{itemize}

\subsection{Implementation Details}
TM-HGNN is implemented by PyTorch \cite{paszke2019pytorch} and optimized with Adam \cite{kingma2015adam} optimizer with learning rate 0.001 and dropout rate 0.3. We set hidden dimension $d$ of each layer to 64 and batch size to 32 by searching parameters. We train models for 100 epochs with early-stopping strategy, where the epoch of 30 shows the best results. All experiments are trained on a single NVIDIA GeForce RTX 3080 GPU. 

\section{Results}
Since the dataset has imbalanced class labels for in-hospital mortality as shown in Table \ref{table1}, we use AUPRC (Area Under the Precision-Recall Curve) and AUROC (Area Under the Receiver Operating Characteristic Curve) for precise evaluation. It is suggested by \citet{10.1145/1143844.1143874} to use AUPRC for imbalanced class problems.
\begin{table*}[t]
\centering{
\resizebox{\textwidth}{!}{\begin{tabular}{cccccccccc}
\toprule
\multirow{2}*{\textbf{Categories}} &  \multirow{2}*{\textbf{Models}} & \multicolumn{2}{c}{\textbf{Whole}} 
& \multicolumn{2}{c}{\textbf{Hypertension}} & \multicolumn{2}{c}{\textbf{Diabetes}} \\ 
\cmidrule(lr){3-4}\cmidrule(lr){5-6}\cmidrule(lr){7-8}\cmidrule(lr){9-10}
& & AUPRC & AUROC & AUPRC& AUROC &AUPRC& AUROC \\
\midrule
\multirow{2}*{Word-based} 
& Word2vec + MLP & 13.49 $\pm$ 1.68 & 56.65 $\pm$ 5.12 
& 16.82 $\pm$ 1.78 & 53.56 $\pm$ 4.20 & 18.15 $\pm$ 1.42 & 51.94 $\pm$ 3.40\\
~ & FastText & 17.06 $\pm$ 0.08 & 62.37  $\pm$ 0.11 
& 25.56 $\pm$ 0.28 & 62.39 $\pm$ 0.18 & 31.33 $\pm$ 0.33 & 67.59 $\pm$ 0.20\\ 
\hline
\multirow{3}*{Sequence-based} 
& Bi-LSTM & 17.67 $\pm$ 4.19 & 58.75  $\pm$ 5.78 
& 21.75 $\pm$ 5.25 & 57.39 $\pm$ 6.11 & 27.52 $\pm$ 7.57 & 61.86 $\pm$ 8.38 \\
~ & Bi-LSTM w/ Att. & 17.96 $\pm$ 0.61 & 62.63 $\pm$ 1.31 
& 26.05 $\pm$ 1.80 & 63.24 $\pm$ 1.57 & 33.01 $\pm$ 3.53 & 68.89 $\pm$ 1.58\\
~ & TextCNN & 20.34 $\pm$ 0.67 & 68.25 $\pm$ 0.54 
& 27.10 $\pm$ 1.82 & 66.10 $\pm$ 1.20 & 36.89 $\pm$ 2.54 & 71.83 $\pm$ 1.69\\
\hline
\multirow{2}*{Graph-based} 
& TextING & 34.50 $\pm$ 7.79 & 78.20 $\pm$ 4.27 
& 36.63 $\pm$ 8.30 & 80.12 $\pm$ 4.05 & 36.13 $\pm$ 8.66 & 80.28 $\pm$ 3.84\\
~ & InducT-GCN & 43.03 $\pm$ 1.96 & 82.23 $\pm$ 0.72 
& 41.06 $\pm$ 2.95 & 85.56 $\pm$ 1.24 & 40.59 $\pm$ 3.07 & 84.42 $\pm$ 1.45\\
\hline
\multirow{3}*{HyperGraph-based} 
& HyperGAT & 44.42 $\pm$ 1.96 & 84.00 $\pm$ 0.84 
& 42.32 $\pm$ 1.78 & 86.41 $\pm$ 1.01 & 40.08 $\pm$ 2.45 & 85.03 $\pm$ 1.20\\
~ & T-HGNN (Ours) & 45.85 $\pm$ 1.91 & 84.29 $\pm$ 0.31 
& 43.53 $\pm$ 2.01 & 87.07 $\pm$ 0.64 & 40.47 $\pm$ 2.29 & 85.48 $\pm$ 0.92\\
~ & TM-HGNN (Ours) & \textbf{48.74 $\pm$ 0.60} & \textbf{84.89 $\pm$ 0.42} 
& \textbf{47.27} $\pm$ \textbf{1.21} & \textbf{87.75} $\pm$ \textbf{0.54} & \textbf{42.22} $\pm$ \textbf{1.25} & \textbf{85.86} $\pm$ \textbf{0.73}\\
\bottomrule
\end{tabular}}}
\caption{Classification performance comparison on patient-level clinical tasks, evaluated with AUPRC and AUROC in percentages. We report averaged results with standard deviation over 10 random seeds. Values in boldface denote the best results.}
\label{table2}
\end{table*}
\subsection{Classification Performance}
Table \ref{table2} shows performance comparisons of TM-HGNN and baseline methods. Sequence-based methods outperform word-based methods, which indicates capturing local dependencies between neighboring words benefits patient document classification. Moreover, all graph-based methods outperform sequence-based and word-based methods. This demonstrates ignoring sequential information of words is not detrimental to clinical notes. Furthermore, hypergraphs are more effective than previous word co-occurrence graphs, indicating that it is crucial to extract high-order relations within clinical notes. In particular, as TM-HGNN outperforms HyperGAT \cite{ding2020more}, exploiting taxonomy-level semantic information which represents the medical context of the notes aids precise prediction in patient-level. Another advantage of our model, which captures multi-level high order relations from note-level and taxonomy-level with hierarchy, can be verified by the results in Table \ref{table2} where TM-HGNN outperforms T-HGNN. T-HGNN indicates the variant of TM-HGNN, which considers note-level and taxonomy-level hyperedges homogeneous. Likewise, results from hypertension and diabetes patient groups show similar tendencies in overall.   

\begin{figure}[t]
\centering
\includegraphics[width=1\columnwidth]{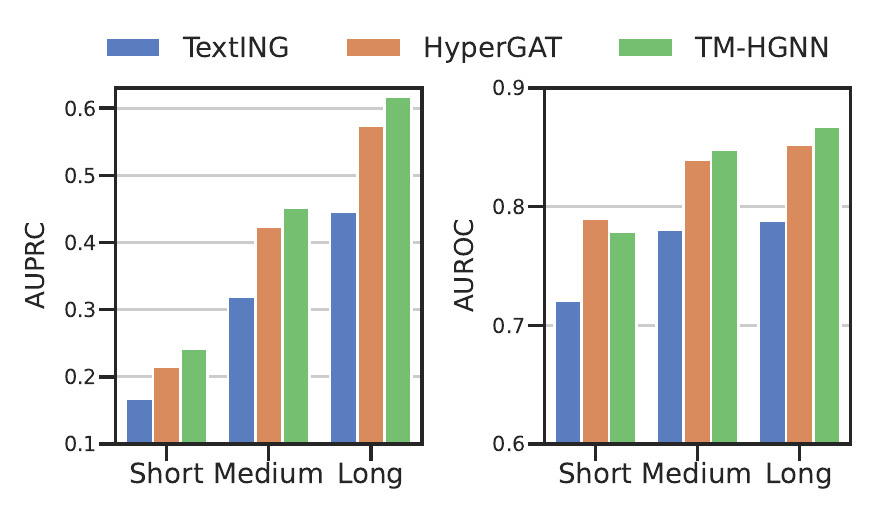} 
\caption{Prediction results of TextING, HyperGAT, and TM-HGNN for three patient-level clinical note groups divided by length (short, medium, and long). AUPRC and AUROC are used for evaluation.}
\label{length_performance}
\end{figure}

\begin{figure}[t]
\centering
\includegraphics[width=1\columnwidth]{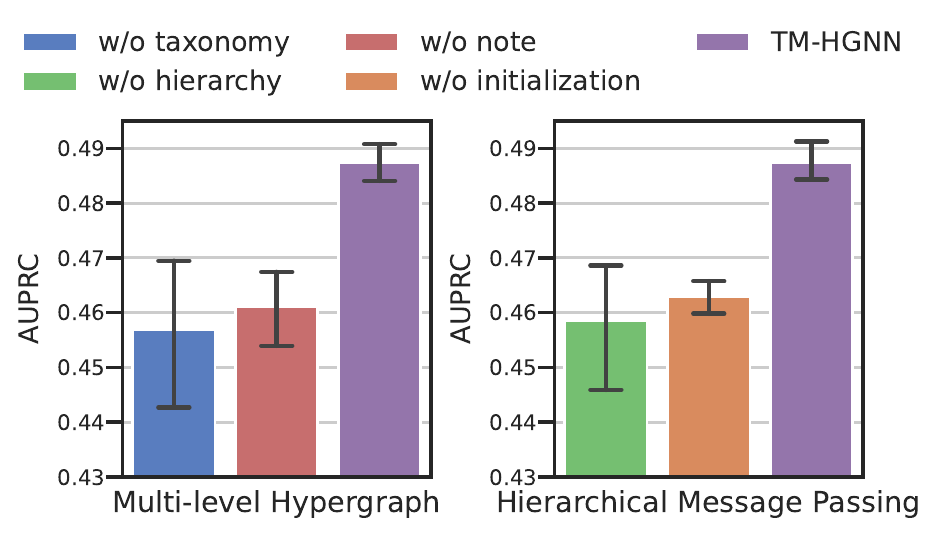} 
\caption{Performance results of ablation studies. The effectiveness of the multi-level hypergraph and hierarchical message passing in the proposed model TM-HGNN are validated respectively.}
\label{ablation}
\end{figure}

\subsection{Robustness to Lengths}
To evaluate the performance dependencies to lengths, we divide clinical notes in patient-level into three groups by lengths, which are short, medium, and long (Appendix \ref{train_stats}, Figure \ref{num_words}). For test set, the number of patients is 645, 1,707, and 856 for short, medium, and long group each, and the percentage of mortality is 6.98\%, 10.72\%, and 15.89\% for each group, which implies patients in critical condition during ICU stays are more likely to have long clinical notes. Figure \ref{length_performance} shows performance comparisons for three divided groups with TextING \cite{zhang2020every} which utilizes word co-occurrence graph, HyperGAT \cite{ding2020more}, a ordinary hypergraph based approach, and our multi-level hypergraph approach (TM-HGNN). All three models were more effective to longer clinical notes, which demonstrates graph based models are robust to long document in general. Among the three models, our proposed TM-HGNN mostly performs the best and HyperGAT \cite{ding2020more} follows, and then TextING \cite{zhang2020every}. The results demonstrate that our TM-HGNN, which exploits taxonomy-level semantic information, is most effective for clinical notes regardless of the lengths, compared to other graph-based approaches. 


\begin{figure*}[t]
\centering
\includegraphics[width=0.95\textwidth]{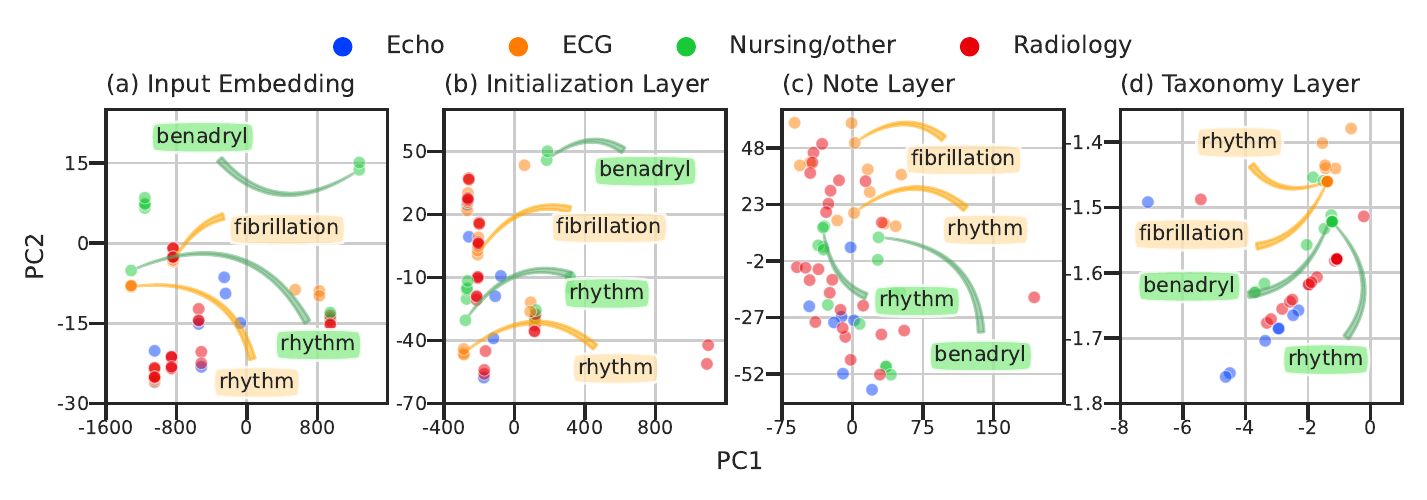} 
\caption{PCA results of learned node representations from each layer of TM-HGNN, for patient case HADM\_ID=147702. "\textit{Rhythm}" and "\textit{fibrillation}" from ECG, "\textit{rhythm}" and "\textit{benadryl}" from Nursing/other taxonomy are highlighted. (a) Input word node embeddings. (b) Initialized node embeddings from the first layer. (c) After second layer, note-level message passing. (d) Final node embeddings from TM-HGNN, after taxonomy-level message passing. Word node embeddings are aligned with the same taxonomy words.}
\label{hierarchical_case}
\end{figure*}

\begin{figure}[t]
\centering
\includegraphics[width=1\columnwidth]{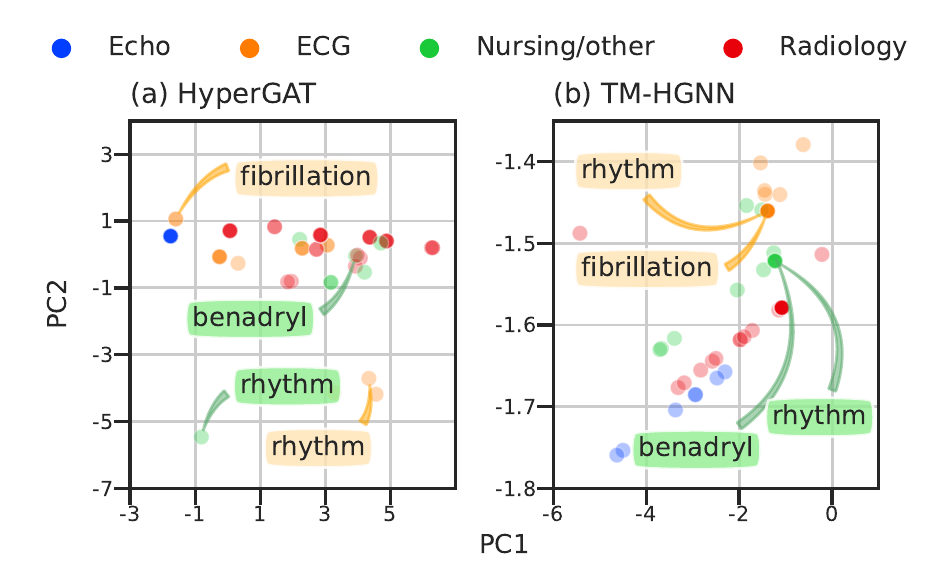} 
\caption{PCA results of learned node representations from HyperGAT (a) and TM-HGNN (b). "\textit{Rhythm}" and "\textit{fibrillation}" from ECG, "\textit{rhythm}" and "\textit{benadryl}" from Nursing/other taxonomy are highlighted.}
\label{taxonomy_case}
\end{figure}

\subsection{Ablation Study}
\paragraph{Effect of Multi-level Hypergraph}
In order to validate the effect of multi-level hypergraphs, we ignore taxonomy-level and note-level hyperedges respectively. \textit{w/o taxonomy}, which ignores taxonomy-level hyperedges, deteriorates the performance most significantly. \textit{w/o note} shows degraded performance as well. Thus, effectiveness of multi-level hypergraph construction for patient representation learning can be verified (Figure \ref{ablation}).
\paragraph{Effect of Hierarchical Message Passing}
Figure \ref{ablation} demonstrates that hierarchical message passing (note-level to taxonomy-level) for multi-level hypergraphs is effective than learning without hierarchies, since  \textit{w/o hierarchy} shows inferior performance compared to TM-HGNN. \textit{w/o hierarchy} represents T-HGNN from Table \ref{table2}, which considers every hyperedge as homogeneous. Degraded performance from \textit{w/o initialization} shows the effectiveness of the initialization layer before hierarchical message passing, which indicates that pre-training on the entire multi-level hypergraphs first benefits the patient-level representation learning.  

\subsection{Case Study}
\paragraph{Hierarchical Message Passing}
We visualize the learned node representations based on principal component analysis (PCA) \cite{jolliffe2002principal} results, as hierarchical message passing continues in TM-HGNN. In Figure \ref{hierarchical_case}(a), "\textit{rhythm}" from ECG and Nursing/other taxonomy are mapped closely for initial word embeddings, since they are literally same words. As the patient-level hypergraphs are fed into a global-level, note-level, and taxonomy-level convolutional layers in order, words in the same taxonomies assemble, which can be found in Figure \ref{hierarchical_case}(b), (c), and (d). As a result, "\textit{rhythm}" of ECG represents different semantic meanings from "\textit{rhythm}" of Nursing/other, as it is learned considerably close to "\textit{fibrillation}" from the same taxonomy. 
\paragraph{Importance of Taxonomy-level Semantic Information} To investigate the importance of taxonomy-level semantic information extraction, we visualize PCA results of the learned node embeddings from the baseline method and the proposed TM-HGNN. We select patient with hospital admission id (HADM\_ID) 147702 for case study since TM-HGNN successfully predicts the true label for in-hospital-mortality, which is positive, but the other baseline methods show false negative predictions. As in Figure \ref{taxonomy_case}, HyperGAT learns "\textit{rhythm}" without taxonomy-level semantic information, since it is not assembled with other words in the same taxonomy. But TM-HGNN separately learns "\textit{rhythm}" from ECG and "\textit{rhythm}" from Nursing/other based on different contexts, which results in same taxonomy words aligned adjacently, such as "\textit{fibrillation}" of ECG and "\textit{benadryl}" of Nursing/other. 
Therefore, in case of TM-HGNN, frequently used neutral word "\textit{rhythm}" from ECG with a word "\textit{fibrillation}" means an irregular "\textit{rhythm}" of the heart and is closely related to mortality of the patient, but "\textit{rhythm}" from Nursing/other with another nursing term remains more neutral. This phenomenon demonstrates that contextualizing taxonomy to frequent neutral words enables differentiation and reduces ambiguity of the frequent neutral words (e.g. "\textit{rhythm}"), which is crucial to avoid false negative predictions on patient-level representation learning.

\section{Conclusion}
In this paper, we propose a taxonomy-aware multi-level hypergraph neural networks, TM-HGNN, a novel approach for patient-level clinical note representation learning. We employ hypergraph-based approach and introduce multi-level hyperedges (note and taxonomy-level) to address long and complex information of clinical notes. TM-HGNN aims to extract high-order semantic information from the multi-level patient hypergraphs in hierarchical order, note-level and then taxonomy-level. Clinical note representations can be effectively learned in an end-to-end manner with TM-HGNN, which is validated from extensive experiments.  \\
\section*{Limitations}
Since our approach, TM-HGNN, aggregates every note during ICU stays for patient representation learning, it is inappropriate for time-series prediction tasks (e.g. vital signs). We look forward to further study that adopts and applies our approach to time-series prediction tasks.

\section*{Ethics Statement}
In MIMIC-III dataset \cite{johnson2016mimic}, every patient is deidentified, according to Health Insurance Portability and Accountability Act (HIPAA) standards. The fields of data which can identify the patient, such as patient name and address, are completely removed based on the identifying data list provided in HIPAA. In addition, the dates for ICU stays are shifted for randomly selected patients, preserving the intervals within data collected from each patient. Therefore, the personal information for the patients used in this study is strictly kept private. More detailed information about deidentification of MIMIC-III can be found in \citet{johnson2016mimic}.

\section*{Acknowledgements}
This work was supported by Institute of Information \& communications Technology Planning \& Evaluation (IITP) grant funded by the Korea government (MSIT) [NO.2021-0-01343, Artificial Intelligence Graduate School Program (Seoul National University)] and the Bio \& Medical Technology Development Program of the National Research Foundation (NRF) funded by the Ministry of Science \& ICT (RS-2023-00257479), and the ICT at Seoul National University provides research facilities for this study.


\bibliography{anthology,custom}
\bibliographystyle{acl_natbib}

\newpage
\appendix

\section{Detailed Statistics of MIMIC-III Clinical Notes }
Table \ref{tax_stat} shows the number of clinical notes assigned to 15 predefined taxonomies in MIMIC-III dataset. Since the number of notes varies in a wide range for each taxonomy, we select top six taxonomies for experiments:  Radiology, ECG, Nursing/other, Echo, Nursing, and Physician.

\begin{figure}[t]
\centering
\includegraphics[width=1\columnwidth]{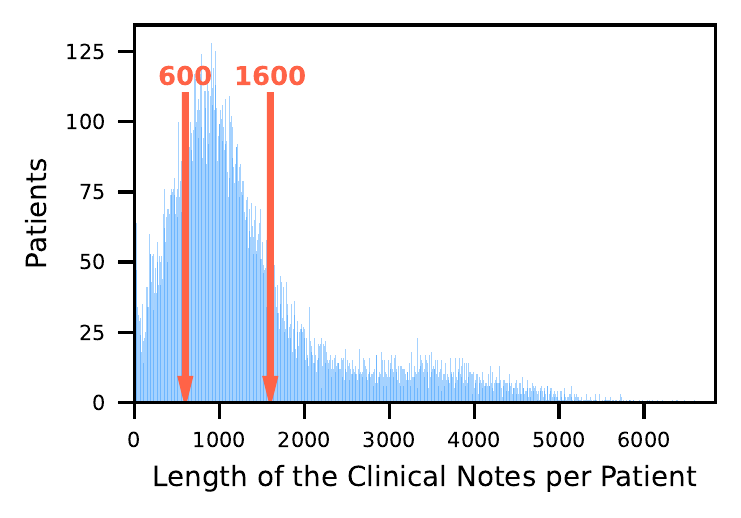} 
\caption{Histogram for the length of patient-level clinical notes in train set. 600 and 1,600 are selected as boundaries to divide clinical notes into three groups (short, medium, and long).}
\label{num_words}
\end{figure}

Figure \ref{num_words} shows histogram for the number of words per patient-level clinical notes in train set. Since 682, 1,070, and 1,689 are the first, second, and third quantile of the train data, we select 600 and 1,600 as the boundaries to divide test set into 3 groups (short, medium, and long), which is used to validate proposed TM-HGNN's robustness to lengths.

\section{Node Representations from Other Methods}
\begin{figure*}[t]
\centering
\includegraphics[width=0.8\textwidth]{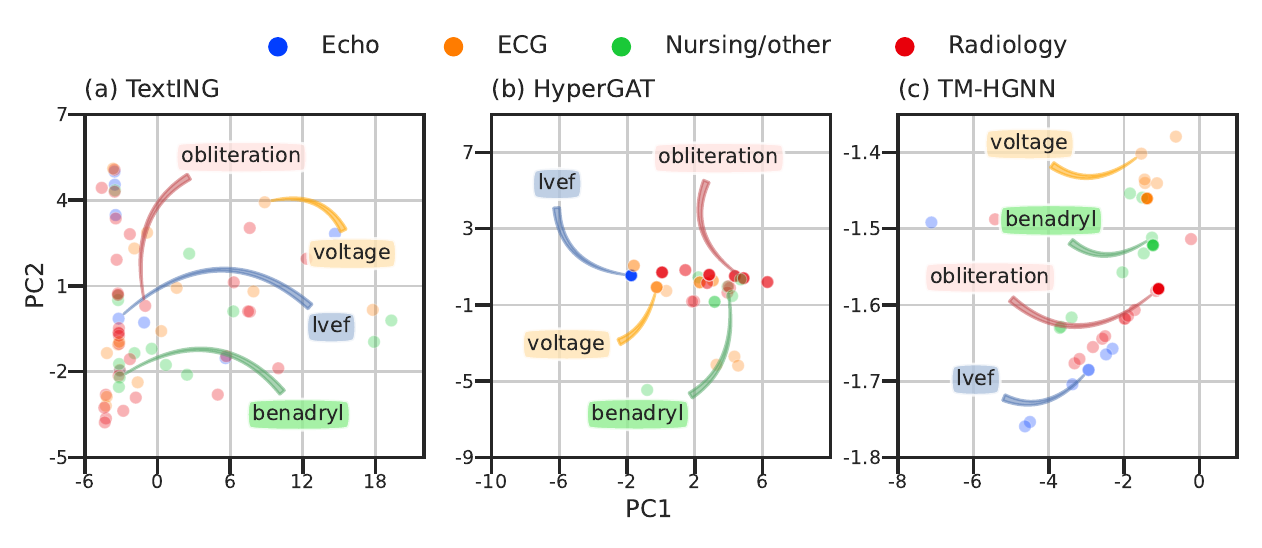} 
\caption{PCA results of learned node representations for patient case HADM\_ID=147702, compared with baseline methods. (a) Final node embeddings from TextING. (b) Final node embeddings from HyperGAT. (c) Final node embeddings from the proposed TM-HGNN.}
\label{case_study}
\end{figure*}

Figure \ref{case_study} shows PCA results of learned node representations from three different models. According to Figure \ref{case_study}(a) and \ref{case_study}(b), word co-occurrence graphs (TextING) and homogeneous single-level hypergraphs (HyperGAT) show node representations ambiguous to discriminate by taxonomies, since every taxonomy has been shuffled. In Figure \ref{case_study}(c), node embeddings are aligned adjacently and arranged with similar pattern for the same taxonomies. This verifies the effectiveness of the proposed TM-HGNN which captures intra- and inter-taxonomy semantic word relations for patient-level representation learning. Example words (voltage, lvef, benadryl, and obliteration) which are generally used in each taxonomy are shown in Figure \ref{case_study} to emphasize that the keywords from each taxonomy are learned adjacently to words similar in context within taxonomies in case of TM-HGNN, but not for other methods. 

\label{train_stats}

\begin{table}[t]
\centering
\resizebox{0.65\columnwidth}{!}{\begin{tabular}{lll}
\toprule
& \# of Notes \\
\hline
Radiology & 17,466 \\
ECG & 16,410 \\
Nursing/other & 12,347\\
Echo & 7,935\\
Nursing & 3,562 \\
Physician & 3,545 \\
Respiratory & 2,024 \\
Nutrition & 1,270 \\
General & 1,135 \\
Discharge Summary & 608 \\
Rehab Services & 594 \\
Social Work & 424 \\
Case Management & 162 \\
Consult & 19 \\
Pharmacy & 14\\
\bottomrule
\end{tabular}}
\caption{The number of clinical notes for 15 predefined taxonomies in MIMIC-III dataset.}
\label{tax_stat}
\end{table}

\section{Explanation of the Medical Terms}
\begin{itemize}
\item Fibrillation : Fibrillation refers to rapid and irregular contractions of the muscle fibers, especially from the heart. It can lead to serious heart conditions.
\item Benadryl : Brand name for the drug Diphenhydramine, which is an antihistamine. Benadryl is one of the over-the-counter drugs, and generally used for alleviating the allergic symptoms. 
\item Lvef : Abbreviation of left ventricular ejection fraction, which is the ratio of stroke volume to end-diastolic volume. Lvef is known as the central measure for the diagnosis and management of heart failure.
\item Obliteration : In Radiology, obliteration refers to the disappearance of the contour of an organ, due to the same x-ray absorption from the adjacent tissue.
\end{itemize}

\section{Additional Performance Comparison}
We conduct additional experiments using LSTM based on 17 code features selected by \citet{johnson2016mimic}, and Transformer-based ClinicalXLNet \cite{huang2020clinical} without pre-training for in-hospital mortality prediction. Table \ref{add_results} shows that TM-HGNN outperforms approaches using structured data and Transformer-based model without pre-training. 

In addition, we train our model on acute kidney injury prediction task (MIMIC-AKI) following \citet{li2023comparative}. Table \ref{mimic-aki} shows comparative results of our TM-HGNN to Clinical-Longformer \cite{li2023comparative} that justify TM-HGNN can effectively utilize high-order semantics from long clinical notes, with much less computational burden compared to long sequence transformer models.

\begin{table}[t]
\centering
\resizebox{1\columnwidth}{!}{
\small
\begin{tabular}{ccc}
\hline
Models & AUPRC & AUROC \\
\hline
LSTM (code features) & 39.86 & 81.98 \\
ClinicalXLNet (w/o pretrain) & 16.77 & 62.16 \\
TM-HGNN (Ours) & \textbf{48.74} & \textbf{84.89} \\
\hline
\end{tabular}}
\caption{Classification performance comparison on patient-level in-hospital-mortality prediction task, evaluated with AUPRC and AUROC in percentages. Values in boldface denote the best results.}
\label{add_results}
\end{table}

\begin{table}[t]
\centering
\resizebox{1\columnwidth}{!}{
\tiny
\begin{tabular}{ccc}
\hline
Models & AUROC & F1 \\
\hline
Clinical-Longformer & 0.762 & \textbf{0.484} \\
TM-HGNN (Ours) & \textbf{0.847} & 0.462 \\
\hline
\end{tabular} }
\caption{Classification performance comparison on patient-level acute kidney injury prediction task, evaluated with AUROC and F1 score. Values in boldface denote the best results.}
\label{mimic-aki}
\end{table}

\end{document}